\documentclass{ecai}
\usepackage{times}
\usepackage{graphicx}
\usepackage{latexsym}

\usepackage[numbers,sort&compress]{natbib}
\usepackage[skip=0pt]{caption}
\usepackage[inline]{enumitem}
\usepackage{multirow}
\usepackage{booktabs} 
\usepackage{xcolor} 
\usepackage{amsmath,bm} 
\usepackage{amssymb} 
\usepackage{acronym} 
\usepackage{hyperref} 
\usepackage[compact]{titlesec}
\usepackage{acronym}
\acrodef{MoGNet}{mixture-of-generators network}
\acrodef{TDS}{task-oriented dialogue system}
\acrodef{DRG}{dialogue response generation}
\acrodef{DCTG}{Dialogue-Context-to-Text Generation}
\acrodef{NLG}{Natural Language Generation}
\acrodef{MDRG}{Modular Dialogue Response Generation}
\acrodef{DRGor}{Dialogue Response Generator}
\acrodef{MTDS}{Modular Task-oriented Dialogue System}
\acrodef{MoE}{Mixture-of-Experts}
\acrodef{RMoG}{retrospective mixture-of-generators}
\acrodef{PMoG}{prospective mixture-of-generators}
\acrodef{RPMoG}{Retrospective and Prospective Mixture-of-Generators}
\acrodef{NMOE}{Neural Mixture-of-Experts}
\acrodef{KB}{Knowledge Base}
\acrodef{SeqMoE}{Sequence-level Mixture-of-Experts}
\acrodef{TokenMoE}{Token-level Mixture-of-Experts}
\acrodef{SentenceMoE}{Sentence-level Mixture-of-Experts}
\acrodef{MLP}{Multilayer Perceptron}
\acrodef{RNN}{Recurrent Neural Network}
\acrodef{GRU}{Gated Recurrent Unit}
\acrodef{Seq2Seq}{Sequence-to-Sequence}
\acrodef{Seq2SeqAttn}{Sequence-to-Sequence with Attention}
\acrodef{LSTM}{Long Short-Term Memory}
\acrodef{S2SAttnLSTM}{Sequence-to-Sequence with Attention Using LSTM}
\acrodef{S2SAttnGRU}{Sequence-to-Sequence with Attention Using GRU}
\acrodef{LaRLAttnGRU}{Latent Action Reinforcement Learning with Attention Using GRU}
\acrodef{GL}{global-and-local}
\acrodef{MultiWOZ}{Multi-domain Wizard-of-Oz}

\acrodef{NLU}{Natural Language Understanding}
\acrodef{DST}{Dialogue State Tracking}
\acrodef{PL}{Policy Learning}
\acrodef{NLG}{Natural Language Generation}

\acrodef{POMDP}{Partially Observable Markov Decision Process}
\acrodef{AMT}{Amazon Mechanical Turk}

\usepackage{amsmath}

\DeclareMathOperator{\softmax}{softmax}
\DeclareMathOperator{\MLP}{MLP}

\begin{document}

\title{Retrospective and Prospective Mixture-of-Generators \\for Task-oriented Dialogue Response Generation}

\author{Jiahuan Pei \and Pengjie Ren \and Christof Monz \and Maarten de Rijke\institute{University of Amsterdam,
The Netherlands, email: \{j.pei, p.ren, c.monz, derijke\}@uva.nl} }

\maketitle

\begin{abstract}
Dialogue response generation~(\acs{DRG})\acused{DRG} is a critical component of task-oriented dialogue systems~(\acsp{TDS})\acused{TDS}.
Its purpose is to generate proper natural language responses given some context, e.g., historical utterances, system states, etc.
State-of-the-art work focuses on how to better tackle \acs{DRG} in an end-to-end way.
Typically, such studies assume that each token is drawn from a single distribution over the output vocabulary, which may not always be optimal. 
Responses vary greatly with different intents, e.g., domains, system actions.
We propose a novel \acf{MoGNet} for \ac{DRG}, where we assume that each token of a response is drawn from a mixture of distributions.
\ac{MoGNet} consists of a chair generator and several expert generators.
Each expert is specialized for \ac{DRG} w.r.t.\ a particular intent. 
The chair coordinates multiple experts and combines the output they have generated to produce more appropriate responses.
We propose two strategies to help the chair make better decisions, namely, a \acf{RMoG} and a \acf{PMoG}.
The former only considers the historical expert-generated responses until the current time step while the latter also considers possible expert-generated responses in the future by encouraging exploration.
In order to differentiate experts, we also devise a \acf{GL} learning scheme that forces each expert to be specialized towards a particular intent using a local loss and trains the chair and all experts to coordinate using a global loss.
We carry out extensive experiments on the \acs{MultiWOZ} benchmark dataset.
\ac{MoGNet} significantly outperforms state-of-the-art methods in terms of both automatic and human evaluations, demonstrating its effectiveness for \acs{DRG}.
\end{abstract}


\section{INTRODUCTION}

\begin{figure}[htb!]
  \begin{minipage}{\columnwidth}
    \centering
    \begin{tabular}{@{}c@{}}
        \includegraphics[width=1\columnwidth, trim=0cm 0cm 0cm 0cm, clip]{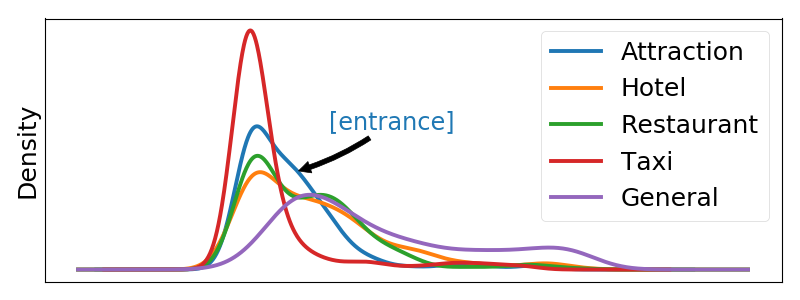} 
        \\
        \includegraphics[width=1\columnwidth, trim=0cm 0cm 0cm 0cm, clip]{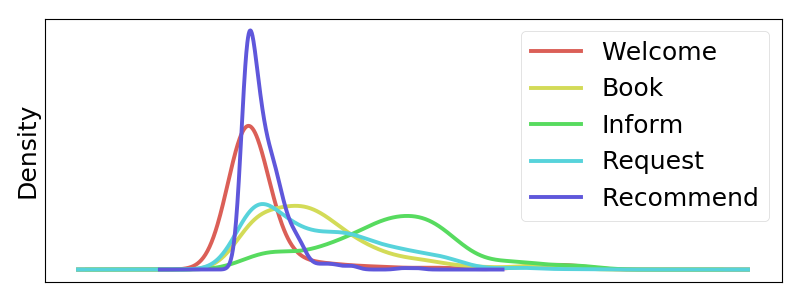} 
    \end{tabular}
    \caption[]{Density of the relative token frequency distribution for different intents (\emph{domains} in the top plot, \emph{system actions} in the bottom plot). We use kernel density estimation\footnotemark\  to estimate the probability density function of a random variable from a relative token frequency distribution.}
    \label{fig:stats}
  \end{minipage}
\end{figure}

\begin{figure*}[htb!]
    \centering
    \includegraphics[width=\textwidth,trim=0 0 50 0,clip]{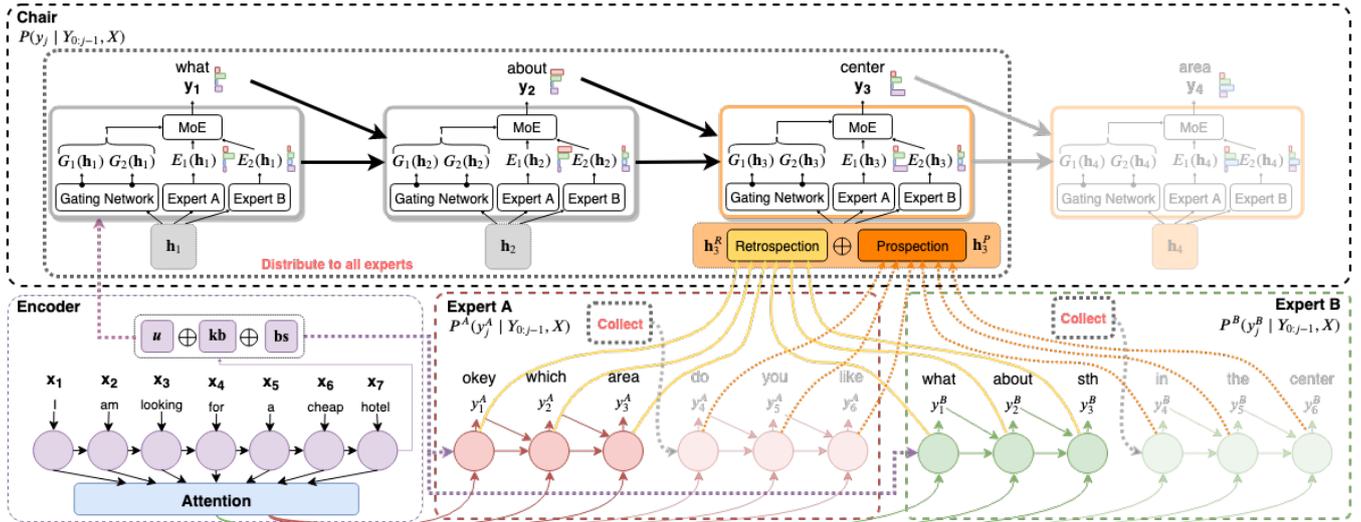}
    \caption{Overview of \ac{MoGNet}. It illustrates how the model generates the token $y_3$ given sequence $X$ as an input in the process of generating the whole sequence $Y$ as a dialogue response.}
    \label{fig:mognet}
\end{figure*}

Task-oriented dialogue systems (\acp{TDS}) have sparked considerable interest due to their broad applicability, e.g., for booking flight tickets or scheduling meetings~\citep{williams2017hybrid,young2013pomdp}.
Existing \ac{TDS} methods can be divided into two broad categories: pipeline multiple-module models~\cite{end2end_dataset_paper_babi_bordes,chen2017survey,young2013pomdp} and end-to-end single-module models~\cite{eric2017key,wen2016network}.
The former decomposes the \ac{TDS} task into sequentially dependent modules that are addressed by separate models while the latter proposes to use an end-to-end model to solve the entire task.
In both categories, there are many factors to consider in order to achieve good performance, such as user intent understanding~\citep{wen2017network}, dialogue state tracking~\citep{zhong2018dst}, and \acf{DRG}.
Given a \textit{dialogue context} (dialogue history, states, retrieved results from a knowledge base, etc.), the purpose of \ac{DRG} is to generate a proper natural language response that leads to task-completion, i.e., successfully achieving specific goals, and that is fluent, i.e., generating natural and fluent utterances.

\footnotetext{\url{https://pandas.pydata.org/pandas-docs/stable/reference/api/pandas.DataFrame.plot.kde.html}}
Recently proposed \ac{DRG} methods have achieved promising results (see, e.g., \acs{LaRLAttnGRU}~\citep{zhao-etal-2019-rethinking}).
However, when generating a response, all current models assume that each token is drawn from a single distribution over the output vocabulary.
This may be unreasonable because responses vary greatly with different intents, where intent may refer to domain, system action, or other criteria for partioning responses, e.g., the source of dialogue context~\cite{pei-2019-sentnet}.
To support this claim, consider the training set of the \acf{MultiWOZ} benchmark dataset~\citep{budzianowski2018multiwoz}, where 67.4\% of the dialogues span across multiple domains and all of the dialogues span across multiple types of system actions.
We plot the density of the relative token frequency distributions in responses of different intents over the output vocabulary in Fig.~\ref{fig:stats}.
Although there is some overlap among distributions, there are also clear differences.
For example, when generating the token $[entrance]$, it has a high probability of being drawn from the distributions for the intent of \emph{booking an attraction}, but not from \emph{booking a taxi}.
Thus, we hypothesize that a response should be drawn from a mixture of distributions for multiple intents rather than from a single distribution for a general intent. 

We propose a \acfi{MoGNet} for \ac{DRG}, which consists of a \emph{chair} generator and several \emph{expert} generators.
Each expert is specialized for a particular intent, e.g., one domain, or one type of action of a system, etc. 
The chair coordinates multiple experts and generates the final response by taking the utterances generated by the experts into consideration.
Compared with previous methods, the advantages of \ac{MoGNet} are at least two-fold: 
First, the specialization of different experts and the use of a chair for combining the outputs breaks the bottleneck of a single model~\citep{dietterich2000ensemble,masoudnia2014mixture}.
Second, it is more easily \emph{traceable}: we can analyze who is responsible when the model makes a mistake and generates an inappropriate response.

We propose two strategies to help the chair make good decisions, i.e., \acfi{RMoG} and \acfi{PMoG}.
\ac{RMoG} only considers the retrospective utterances generated by the experts, i.e.,  the utterances generated by all the experts prior to the current time step.
However, a chair without a long-range vision is likely to make sub-optimal decisions.
Consider, for example, these two responses: ``what {\bf day} will you be traveling?'' and ``what {\bf day} and {\bf time} would you like to travel?'' 
If we only consider these responses until the 2nd token (which \ac{RMoG} does), then the chair might choose the first response due to the absence of a more long-range view of the important token ``time'' located after the 2nd token. 
Hence, we also propose a \ac{PMoG}, which enables the chair to make full use of the prospective predictions of experts as well.

To effectively train \ac{MoGNet}, we devise a \acfi{GL} learning scheme.
The local loss is defined on a segment of data with a certain intent, which forces each expert to specialize.
The global loss is defined on all data, which forces the chair and all experts to coordinate with each other.
The global loss can also improve data utilization by enabling the backpropagation error of each data sample to influence all experts as well as the chair. 

To verify the effectiveness of \ac{MoGNet}, we carry out experiments on the \acs{MultiWOZ} benchmark dataset.
\ac{MoGNet} significantly outperforms state-of-the-art \ac{DRG} methods, improving over the best performing model on this dataset by 5.64\% in terms of overall performance (0.5*\textit{Inform}$+$0.5*\textit{Success}$+$\textit{BLEU}) and 0.97\% in terms of response generation quality (\textit{Perplexity}).

The main contributions of this paper are:
\begin{itemize}[nosep]
\item a novel \ac{MoGNet} model that is the first framework that devises chair and expert generators for \ac{DRG}, to the best of our knowledge;
\item two novel coordination mechanisms, i.e., \ac{RMoG} and \ac{PMoG}, to help the chair make better decisions; and
\item a \ac{GL} learning scheme to differentiate experts and fuse data efficiently.
\end{itemize}


\section{MIXTURE-OF-GENERATORS NETWORK}
\label{sec:tmoe}

We focus on task-oriented \acs{DRG} (a.k.a. the context-to-text generation task \citep{budzianowski2018multiwoz}).
Formally, given a current dialogue context $X = (U, B, D)$, where $U$ is a combination of previous utterances, $B$ are the belief states, and $D$ are the retrieved database results based on $B$, the goal of task-oriented \acs{DRG} is to generate a fluent natural language response $Y = (y_1, \dots, y_n)$ that contains appropriate system actions to help users accomplish their task goals, e.g., booking a flight ticket.
We propose \ac{MoGNet} to model the generation probability $P(Y\mid X)$.

\subsection{Overview}
The \acs{MoGNet} framework consists of two types of roles:
\begin{itemize}[nosep]
    \item $k$ \textbf{expert generators}, each of which is specialized for a particular \textit{intent}, e.g., a domain, a type of action of a system, etc.
    Let $\mathcal{D}=\{(X_p, Y_p)\}_{p=1}^{|\mathcal{D}|}$ denote a dataset with $|\mathcal{D}|$ independent samples of $(X, Y)$.
    Expert-related intents partition $\mathcal{D}$ into $k$ pieces $\mathcal{S}=\{\mathcal{S}_l\}_{l=1}^{k}$, where $\mathcal{S}_l \triangleq \{(X^l_p, Y^l_p)\}_{p=1}^{|\mathcal{S}_l|}$.
    Then $\mathcal{S}_l$ is used to train each expert by predicting $P^l (Y^l\mid X^l)$.
    We expect the $l$-{th} expert to perform better than the others on $\mathcal{S}_l$.

    \item a \textbf{chair generator}, which learns to coordinate a group of experts to make an optimal decision.
    The chair is trained to predict $P(Y\mid X)$, where $(X, Y)$ is a sample from $\mathcal{D}$.
\end{itemize}

\noindent%
Fig.~\ref{fig:mognet} shows our implementation of $\acs{MoGNet}$; it consists of three types of components, i.e., a shared context encoder, $k$ expert decoders, and a chair decoder.

\subsection{Shared context encoder}
The role of the shared context encoder is to read the dialogue context $X$ and construct a representation.
We follow \citet{budzianowski2018towards} and model the current dialogue context as a combination of user utterances $U$, belief states $B$, and retrieval results from a database $D$. 

First, we employ a \ac{RNN} \citep{cho2014rnn} to map a sequence of input tokens $U=\{w_1, \ldots, w_m\}$ to hidden vectors $\mathbf{H}^{U}=\{\mathbf{h}^{U}_1, \dots, \mathbf{h}^{U}_m\}$.
The hidden vector $\mathbf{h}_i$ at the $i$-{th} step can be represented as:
\begin{equation}
\label{encoder_rnn}
    \mathbf{h}^{U}_i, \mathbf{s}_i =\text{RNN}(\mathbf{w}_i, \mathbf{h}^{U}_{i-1}, \mathbf{s}_{i-1}),
\end{equation}
where $\mathbf{w}_i$ is the embedding of the token $w_i$. 
The initial state $\mathbf{s}_{0}$ of the \acs{RNN} is set to 0.

Then, we represent the current dialogue context $\mathbf{x}$ as a combination of the user utterance representation $\mathbf{h}^{U}_m$, the belief state vector $\mathbf{h}^{B}$, and the database vector $\mathbf{h}^{D}$:
\begin{equation}
    \mathbf{x} = \tanh(\mathbf{W}_u \mathbf{h}^{U}_m + \mathbf{W}_{b} \mathbf{h}^{B} + \mathbf{W}_{d} \mathbf{h}^{D}),
\end{equation}
where $\mathbf{h}^{U}_m$ is the final hidden state from Eq.~\ref{encoder_rnn};
$\mathbf{h}^{B}$ is a 0-1 vector with each dimension representing a state (slot-value pair);
$\mathbf{h}^{D}$ is also a 0-1 vector, which is built by querying the database with the current state $B$.
Each dimension of $\mathbf{h}^{D}$ represents a particular result from the database (e.g., whether a flight ticket is available).

\subsection{Expert decoder}
Given the current dialogue context $X$ and the current decoded tokens $Y_{0:j-1}$, the $l$-{th} expert outputs the probability $P^l(y_j^l\mid Y_{0:j-1}, X)$ over the vocabulary $\mathcal{V}$ at the $j$-{th} step by:
\begin{equation}\label{expertp}
\begin{split}
P^l(y_j^l\mid Y_{0:j-1}, X) &= \softmax(\mathbf{U}^{T}\mathbf{o}_j^l+\mathbf{b}) \\
\mathbf{o}_j^l, \mathbf{s}_j^l &= \text{RNN}(\mathbf{y}_{j-1} \oplus \mathbf{c}^{l}_j, \mathbf{o}_{j-1}^l, \mathbf{s}_{j-1}^l),
\end{split}
\end{equation}
where $\mathbf{U}$ is the parameter matrix and $\mathbf{b}$ is bias;
$\mathbf{s}_j^l$ is the state vector, which is initialized by the dialogue context vector from the shared context encoder, i.e., $\mathbf{s}_0^l=\mathbf{x}$;
$\mathbf{y}_{j-1}$ is the embedding of the generated token at time step $j-1$;
$\oplus$ is the concatenation operation;
$\mathbf{c}^l_j$ is the context vector which is calculated with a concatenation attention mechanism \citep{bahdanau2014neural,D15-1166} over the hidden representations from a shared context encoder as follows:
\begin{equation}\label{expertc}
\begin{split}
\mathbf{c}^{l}_j & = \sum_{i=1}^{m}\alpha^{l}_{ji}\mathbf{h}_i \\
\alpha^{l}_{ji} & = \frac{\exp(w^{l}_{ji})}{\sum_{i=1}^{m} \exp({w^{l}_{ji})}} \\
w^{l}_{ji} & = \mathbf{v}_l^T\tanh{(\mathbf{W}_l^T(\mathbf{h}_i \oplus \mathbf{s}^{l}_{j-1})+\mathbf{b}_l)},
\end{split}
\end{equation}
where $\alpha$ is a set of attention weights;
$\oplus$ is the concatenation operation.
$\mathbf{W}_l$, $\mathbf{b}_l$, $\mathbf{v}_l$ are learnable parameters, which are not shared by different experts in our experiments.

\subsection{Chair decoder}
Given the current dialogue context $X$ and the current decoded tokens $Y_{0:j-1}$, the chair decoder estimates the final token prediction distribution $P(y_j\mid Y_{0:j-1}, X)$ by combining the prediction probabilities from $k$ experts.
Here, we consider two strategies to leverage the prediction probabilities from experts, i.e., \ac{RMoG} and \ac{PMoG}.
The former only considers expert generator outputs from \textit{history} (until the $(j-1)$-{th} time step), which follows the typical neural \acf{MoE} architecture~\citep{schwab2019granger,shazeer2017outrageously}.
We propose the latter to make the chair generator envision the \textit{future} (i.e., after the $(j-1)$-th time step) by exploring expert generator outputs from $t$  extra steps ($t \in [1, n-j], t \in \mathbb{N}$).

Specifically, the chair determines the prediction $P(y_j\mid Y_{0:j-1}, X)$ as follows:
\begin{equation}\label{chairp}
\begin{split}
\mbox{}\hspace*{-1mm}
P(y_j\mid Y_{0:j-1}, X) = {}
 &\beta_{j}^{C}\cdot P(y_j^c\mid Y_{0:j-1}, X)  \\
 & {}+ \sum_{l=1}^{k} (\beta_{j}^{l,R}+\beta_{j}^{l,P}) \cdot P(y_j^l\mid Y^l_{:j-1}, X),
\end{split}
\end{equation}
where $P(y_j^c\mid Y_{0:j-1}, X)$ is the prediction probability from the chair itself;
$P(y_j^l\mid Y_{0:j-1}, X)$ is the prediction probability from expert $l$;
$\beta_{j}=[\beta_{j}^{C}, \beta_{j}^{l,R}, \beta_{j}^{l,P}]$ are normalized coordination coefficients, which are calculated as:
\begin{equation}
\label{beta}
\begin{split}
\beta_{j} ={}& \frac{\exp(\mathbf{v}^T \mathbf{h}_{j})}{\sum_{l=1}^{k} \exp(\mathbf{v}^T \mathbf{h}_{l})} \\
\mathbf{h}_{j} ={}& \MLP ([P(y_j^c\mid Y_{0:j-1}, X), \mathbf{h}_j^{R}, \mathbf{h}_j^{P}]).
\end{split}
\end{equation}
$\beta_{j}^{C}$, $\beta_{j}^{l,R}$ and $\beta_{j}^{l,P}$ are estimated w.r.t. $P(y_j^c\mid Y_{0:j-1}, X)$, $\mathbf{h}_j^{R}$ and $\mathbf{h}_j^{P}$, respectively.
$\mathbf{h}_j^{R}$ is a list of retrospective decoding outputs from all experts, which is defined as follows:
\begin{equation}
\begin{split}
\mathbf{h}_j^{R} ={}& P(y_{1:j-1}^1\mid y_0, X) \oplus \dots \oplus P(y_{1:j-1}^l\mid y_0, X)\\{}& \oplus P(y_{1:j-1}^k\mid y_0, X),
\end{split}
\end{equation}
where $y_0$ is a special token ``[BOS]'' indicating the start of decoding;
$P(y_{1:j-1}^l\mid y_0, X)$ is the output of expert $l$ from the {1}-st to the $(j-1)$-th step using Eq.~\ref{expertp};
$\mathbf{h}_j^{P}$ is a list of prospective decoding outputs from all experts, which is defined as follows:
\begin{equation}
\begin{split}
\mathbf{h}_j^{P} ={}& P(y_{j:j+t}^1\mid Y_{0:j-1}, X) \oplus \cdots {} \\
& \oplus P(y_{j:j+t}^l\mid Y_{0:j-1}, X)\\
& \oplus P(y_{j:j+t}^k\mid Y_{0:j-1}, X),
\end{split}
\end{equation}
where $P(y_{j:j+t}^l\mid Y_{0:j-1}, X)$ are the outputs of expert $l$ from the $j$-th to ($j+t$)-th step.
We obtain $P(y_{j:j+t}^l\mid X)$ by forcing expert $l$ to generate $t$ steps using Eq.~\ref{expertp} based on the current generated tokens $Y_{0:j-1}$.

\subsection{Learning scheme} 
We devise a global-and-local learning scheme to train \ac{MoGNet}.
Each expert $l$ is optimized by a localized expert loss defined on $\mathcal{S}_l$, which forces each expert to specialize on one of the portions of data $\mathcal{S}_l$.
We use cross-entropy loss for each expert and the joint loss for all experts is as follows:
\begin{equation}
\label{expertloss}
\mathcal{L}_\mathit{experts} = 
 \sum_{l=1}^{k} \sum_{(X^l_p, Y^l_p) \in \mathcal{S}_l} \sum_{j=1}^{n} \mu_{l} y_{j}^{l} \log P(y^l_j\mid Y^l_{0:j-1}, X),
\end{equation}
where $P(y^l_j\mid Y^l_{0:j-1}, X)$ is the token prediction by expert $l$ (Eq.~\ref{expertp}) computed on the $r$-th data sample;
$y_j^{l}$ is a one-hot vector indicating the ground truth token at $j$.

We also design a global chair loss to differentiate the losses incurred from different experts. 
The chair can attribute the source of errors to the expert in charge.
For each data sample in $\mathcal{D}$, we calculate the combined taken prediction $P(y_j\mid Y_{0:j-1}, X)$ (Eq.~\ref{chairp}). 
Then the global loss becomes:
\begin{equation}
\begin{split}
\mathcal{L}_\mathit{chair} 
&=\sum_{r=1}^{|\mathcal{D}|} \sum_{j=1}^{n} y_j \log P(y_j\mid Y_{0:j-1}, X).
\end{split}
\end{equation}
Our overall optimization follows the joint learning paradigm that is defined as a weighted combination of constituent losses:
\begin{equation}\label{finalloss}
\mathcal{L} = \lambda \cdot \mathcal{L}_\mathit{experts} + (1 - \lambda) \cdot \mathcal{L}_\mathit{chair},
\end{equation}
where $\lambda$ is a hyper-parameter to regulate the importance between the experts and the chair for optimizing the loss.


\section{EXPERIMENTAL SETUP}
\label{sec:experimental-setup}

\subsection{Research questions}

We seek to answer the following research questions:
\begin{enumerate*}[label=(RQ\arabic*)]
\item Does \acs{MoGNet} outperform state-of-the-art end-to-end single-module \acs{DRG} models? 
\item How does the choice of a particular coordination mechanism (i.e., \acs{RMoG}, \acs{PMoG}, or neither of the two) affect the performance of \acs{MoGNet}?
\item How does the \acs{GL} learning scheme compare to using the general global learning as a learning scheme?
\end{enumerate*}

\subsection{Dataset}
\label{subsec:datasets}
Our experiments are conducted on the \ac{MultiWOZ}~\cite{budzianowski2018multiwoz} dataset. 
This is the latest large-scale human-to-human \ac{TDS} dataset with rich semantic labels, e.g., domains and dialogue actions, and benchmark results of response generation.\footnote{\url{http://dialogue.mi.eng.cam.ac.uk/index.php/corpus/}}
\ac{MultiWOZ} consists of $\sim$10k natural conversations between a tourist and a clerk. 
It has 6 specific action-related domains, i.e., \textit{Attraction}, \textit{Hotel}, \textit{Restaurant}, \textit{Taxi}, \textit{Train}, and \textit{Booking}, and 1 universal domain, i.e., \textit{General}. 
67.4\% of the dialogues are cross-domain which covers 2--5 domains on average. 
The average number of turns per dialogue is 13.68; a turn contains 13.18 tokens on average.
The dataset is randomly split into into 8,438/1,000/1,000 dialogues for training, validation, and testing, respectively.

\subsection{Model variants and baselines}
We consider a number of variants of the proposed mixture-of-generators model:
\begin{itemize}[nosep]
    \item \textbf{\acs{MoGNet}}: the proposed model with \acs{RMoG} and \acs{PMoG} and \acs{GL} learning scheme.
    \item \textbf{\acs{MoGNet}-P}: the model without prospection ability by removing \acs{PMoG} coordination mechanism from \acs{MoGNet}.
    \item \textbf{\acs{MoGNet}-P-R}: the model removing the two coordination mechanisms and remaining \acs{GL} learning scheme.
    \item \textbf{\acs{MoGNet}-GL}: the model that removes \acs{GL} learning scheme from \acs{MoGNet}.
\end{itemize}
See Table~\ref{tab:abl_setting} for a summary.
Without further indications, the \emph{intents} used are based on identifying eight different domains: Attraction, Booking, Hotel, Restaurant, Taxi, Train, General, and UNK.

\begin{table}
\centering
\setlength{\tabcolsep}{14pt}
\caption{Model variants.}
\label{tab:abl_setting}
\begin{tabular}{lcccc}
\toprule
                          & $\bm{\beta_{j}^{C}}$&  $\bm{\beta_{j}^{l,R}}$ & $\bm{\beta_{j}^{l,P}}$  & $\bm{\lambda}$\\ \midrule
\acs{MoGNet}              & True&True&True             & 0.5 \\
\acs{MoGNet}-P            & True&True&False             & 0.5 \\ 
\acs{MoGNet}-P-R          & True&False&False             & 0.5 \\ 
\acs{MoGNet}-\acs{GL}     & True& True&True             & 0.0 \\ 
\bottomrule 
\end{tabular}
\begin{minipage}{\columnwidth}
\footnotesize{$\beta_{j}^{C}$, $\beta_{j}^{l,R}$, $\beta_{j}^{l,P}$ are from Eq.~\ref{chairp}. ``True'' means we preserve it and learn it as it is. ``False'' means we remove it (set it to 0).
$\lambda$ is from Eq.~\ref{finalloss} and we report two settings, 0.0 and 0.5. See \S~\ref{sec:analysis3}.}
\end{minipage}
\end{table}

\noindent%
To answer RQ1, we compare \ac{MoGNet} with the following methods that have reported results on this task according to the official leaderboard.\footnote{The Context-to-Text Generation task at \url{https://github.com/budzianowski/multiwoz}.\label{multiwoz}}
\begin{itemize}[nosep]
    \item \textbf{\acs{S2SAttnLSTM}}. We follow the dominant \ac{Seq2Seq} model under an encoder-decoder architecture~\cite{chen2017survey} and reproduce the benchmark baseline, i.e., single-module model named \acs{S2SAttnLSTM}~\cite{budzianowski2018multiwoz,budzianowski2018towards}, based on the source code provided by the authors. See footnote~\ref{multiwoz}.
    \item \textbf{\acs{S2SAttnGRU}}. A variant of \acs{S2SAttnLSTM}, with \acp{GRU} instead of LSTMs and other settings kept the same.
    \item \textbf{Structured Fusion}. It learns the traditional dialogue modules and then incorporates these pre-trained sequentially dependent modules into end-to-end dialogue models by structured fusion networks~\cite{mehri2019structured}. 
    \item \textbf{\acs{LaRLAttnGRU}}. The state-of-the-art model~\cite{zhao-etal-2019-rethinking}, which uses reinforcement learning and models system actions as latent variables. \acs{LaRLAttnGRU} uses ground truth system action annotations and user goals to estimate the rewards for reinforcement learning during training.
\end{itemize}

\subsection{Evaluation metrics}
We use the following commonly used evaluation metrics \cite{budzianowski2018multiwoz,zhao-etal-2019-rethinking}:
\begin{itemize}[nosep]
\item \textit{Inform}: the fraction of responses that provide a correct entity out of all responses.
\item \textit{Success}: the fraction of responses that answer all the requested attributes out of all responses.
\item \textit{BLEU}: for comparing the overlap between a generated response to one or more reference responses. 
\item \textit{Score}: defined as  $\textit{Score} = (0.5*\textit{Inform}+ 0.5*\textit{Success}+\textit{BLEU})*100$. This measures the overall performance in term of both task completion and response fluency~\cite{mehri2019structured}.
\item \textit{PPL}: denotes the perplexity of the generated responses, which is defined as the exponentiation of the entropy. This measures how well a probability \acs{DRG} model predicts a token in a response generation process.
\end{itemize}

\noindent%
We use the toolkit released by~\citet{budzianowski2018towards} to compute the metrics.\footnote{\url{https://github.com/budzianowski/multiwoz}.}
Following their settings, we also use \textit{Score} as the selection criterion to choose the best model on the validation set and report the performance of the model on the test set. 
We use a paired t-test to measure statistical significance ($p <0.01$) of relative improvements.

\subsection{Implementation details}
Theoretically, the training time complexity of each data sample is $\mathcal{O}(n*(k+1)*n)$, where $n$ is the number of response tokens.
To reduce the computation cost, we assign $j+t=n$ and compute the expert prediction with Eq.~\ref{expertp}.
This means that the chair will make a final decision only after all the experts have decoded their final tokens. 
Thus, the time complexity decreases to $\mathcal{O}(n*(k+1)+n)$.

For a fair comparison, the vocabulary size is the same as \citet{budzianowski2018multiwoz}, which has 400 tokens.
Out-of-vocabulary words are replaced with ``[UNK]''.
We set the word embedding size to 50 and all \ac{GRU} hidden state sizes to 150.
We use Adam~\cite{adam_optimizer} as our optimization algorithm with hyperparameters $\alpha=0.005$, $\beta_1=0.9$, $\beta_2=0.999$ and $\epsilon= 10^{-8}$. 
We also apply gradient clipping \cite{pmlr-v28-pascanu13} with range [--5, 5] during training.
We use $l2$ regularization to alleviate overfitting, the weight of which is set to $10^{-5}$.
We set the mini-batch size to 64.
We use greedy search to generate the responses during testing.
Please note that if a data point has multiple intents, then we assign it to each corresponding expert, respectively.
The code is available online.\footnote{\url{https://github.com/Jiahuan-Pei/multiwoz-mdrg}}


\section{RESULTS}
\label{sec:results}

\subsection{Automatic evaluation}

We evaluate the overall performance of \acs{MoGNet} and the comparable baselines on the metrics defined in \S 3.4.
The results are shown in Table~\ref{tab:main_result}.
First of all, \acs{MoGNet} outperforms all baselines by a large margin in terms of overall performance metric, i.e., satisfaction \textit{Score}.
\begin{table}
\setlength{\tabcolsep}{7.5pt}
\centering
\caption{Comparison results of \acs{MoGNet} and the baselines.}
\label{tab:main_result}
\begin{tabular}{@{}l@{~}lllll@{}}
\toprule
                            & \bf BLEU        & \bf Inform      & 
                            \bf Success        & \bf Score      & \bf PPL  \\ 
\midrule
\acs{S2SAttnLSTM}           & 18.90\%     & 71.33\%     & 60.96\%        & 85.05      & \textbf{3.98} \\ 
\acs{S2SAttnGRU}            & 18.21\%     & 81.50\%     & 68.80\%        & 93.36      & 4.12 \\ 
Structured Fusion~\cite{mehri2019structured} & 16.34\% & 82.70\% & 72.10\% & 93.74 & -- \\
\acs{LaRLAttnGRU}~\cite{zhao-etal-2019-rethinking}           & 12.80\%     & 82.78\%     &\textbf{79.20\%}& 93.79      & 5.22 \\ \midrule
\acs{MoGNet}     &\textbf{20.13\%}\rlap{$^\ast$}&\textbf{85.30\%}\rlap{$^\ast$}& 73.30\%        &\textbf{99.43}\rlap{$^\ast$}& 4.25 \\ \bottomrule 
\end{tabular}
\begin{minipage}{\columnwidth}
\footnotesize{\textbf{Bold face} indicates leading results. Significant improvements over the best baseline are marked with $^\ast$ (paired t-test, $p < 0.01$).}
\end{minipage}
\end{table}
It significantly outperforms the state-of-the-art baseline \acs{LaRLAttnGRU} by 5.64\% (\textit{Score}) and 0.97 (\textit{PPL}).
Thus, \acs{MoGNet} not only improves the satisfaction of responses but also improves the quality of the language modeling process.
\acs{MoGNet} also achieves more than 6.70\% overall improvement over the benchmark baseline \acs{S2SAttnLSTM} and its variant \acs{S2SAttnGRU}.
This proves the effectiveness of the proposed \acs{MoGNet} model.

Second, \acs{LaRLAttnGRU} achieves the highest performance in terms of \textit{Success}, followed by \acs{MoGNet}.
However, it results in a 7.33\% decrease in \textit{BLEU} and a 2.56\% decrease in \textit{Inform} compared to \acs{MoGNet}. 
Hence, \acs{LaRLAttnGRU} is good at answering all requested attributes but not as good at providing more appropriate entities with high fluency as \acs{MoGNet}.
\acs{LaRLAttnGRU} tends to generate more slot values to increase the probability of answering the requested attributes.
Take an extreme case as an example: if we force a model to generate all tokens with slot values, then it will achieve an extremely high \textit{Success} but a low \textit{BLEU}.

Third, \acs{S2SAttnLSTM} is the worst model in terms of overall performance (\textit{Score}).
But it achieves the best \textit{PPL}.
It tends to generate frequent tokens from the vocabulary which exhibits better language modeling characteristics.
However, it fails to provide useful information (the requested attributes) to meet the user goals.
By contrast, \acs{MoGNet} improves the user satisfaction (i.e., \textit{Score}) greatly and achieves response fluency by taking specialized generations from all experts into account.

\subsection{Human evaluation}

To further understand the results in Table~\ref{tab:main_result}, we conducted a human evaluation of the generated responses from \acs{S2SAttnGRU}, \acs{LaRLAttnGRU}, and \acs{MoGNet}. 
We ask workers on \acf{AMT}\footnote{https://www.mturk.com/} to read the dialogue context, and choose the responses that satisfy the following criteria: 
(i) \textit{Informativeness} measures whether the response provides appropriate information that is requested by the user query. 
No extra inappropriate information is provided.
(ii) \textit{Consistency} measures whether the generated response is semantically aligned with the ground truth response.
(iii) \textit{Satisfactory} measures whether the response has a overall satisfactory performance promising both \textit{Informativeness} and \textit{Consistency}.
As with existing studies~\cite{mehri2019structured}, we sample one hundred context-response pairs to do human evaluation.
Each sample is labeled by three workers.
The workers are asked to choose either all responses that satisfy the specific criteria or the ``NONE'' option, which denotes none of the responses satisfy the criteria.
To make sure that the annotations are of high quality, we calculate the fraction of the responses that satisfy each criterion out of all responses that passes the \textit{golden test}.
That is, we only consider the data from the workers who have chosen the golden response as an answer.

\begin{table}
\setlength{\tabcolsep}{3.5pt}
\caption{Results of human evaluation.}
\label{tab:human_eval}
\centering
\begin{tabular}{@{}lcccccc@{}}
\toprule
  & \multicolumn{2}{c}{\acs{S2SAttnGRU}} & \multicolumn{2}{c}{\acs{LaRLAttnGRU}} & \multicolumn{2}{c}{\acs{MoGNet}}    \\ 
  \cmidrule(r){2-3}
  \cmidrule(r){4-5}
  \cmidrule{6-7}
  & $\geqslant 1$  & $\geqslant 2$ & $\geqslant 1$  & $\geqslant 2$  & $\geqslant 1$ & $\geqslant 2$ \\ \midrule
Informativeness & 56.79\%        & 31.03\%       & 76.54\%        & 44.83\%        & \textbf{80.25\%}       & \textbf{53.45}\%       \\
Consistency & 45.21\%        & 23.53\%       & 71.23\%        & 39.22\%        & \textbf{80.82\%}       & \textbf{50.98\%}       \\
Satisfactory & 26.79\%        & 25.00\%       & 44.64\%        & 21.88\%        & \textbf{60.71\%}       & \textbf{37.50\%}       \\
\bottomrule
\end{tabular}
\begin{minipage}{\columnwidth}
\footnotesize{\textbf{Bold face} indicates the best results. $\geqslant n$ means that at least $n$ \acs{AMT} workers regard it as a good response w.r.t. \textit{Informativeness}, \textit{Consistency} and \textit{Satisfactory}.}
\end{minipage}
\end{table}

The results are displayed in Table~\ref{tab:human_eval}.
\acs{MoGNet} performs better than \acs{S2SAttnGRU} and \acs{LaRLAttnGRU} on \textit{Informativeness} because it frequently outputs responses that provide richer information (compared with \acs{S2SAttnGRU}) and fewer extra inappropriate information (compared with \acs{LaRLAttnGRU}).
\acs{MoGNet} obtains the best results, which means \acs{MoGNet} is able to generate responses that are semantically similar to the golden responses with large overlaps.
The results of \acs{LaRLAttnGRU} outperforms \acs{S2SAttnGRU} in all cases except for \textit{Satisfactory} under the strict condition ($\geqslant 2$).
This reveals that balancing between \textit{Informativeness} and \textit{Consistency} makes it difficult for the mturk workers to assess the overall quality measured by \textit{Satisfactory}.
In this case, \acs{MoGNet} receives the most votes on \textit{Satisfactory} under the strict condition ($\geqslant 2$) as well as the loose condition ($\geqslant 1$).
This shows that the workers consider the responses from \acs{MoGNet} more appropriate than the other two models with a high degree of agreement.
To sum up, \acs{MoGNet} is able to generate user-favored responses in addition to the improvements for automatic metrics.

\subsection{Coordination mechanisms}

In Table~\ref{tab:mognet_ablation} we contrast the effectiveness of different coordination mechanisms.
We can see that \acs{MoGNet}-P loses 4.32\% overall performance with a 0.62\% decrease of \textit{BLEU}, 5.90\% decrease of \textit{Inform} and 1.50\% decrease of \textit{Success}.
This shows that the prospection design of the \acs{PMoG} mechanism is beneficial to both task completion and response fluency.
Especially, most improvements come from providing more correct entities while improving generation fluency.
\acs{MoGNet}-P-R reduces 2.62\% \textit{Score} with 1.97\% lower of \textit{BLEU}, 0.2\% lower of \textit{Inform} and 1.10\% of \textit{Success}.
Thus, the \ac{MoGNet} framework is effective thanks to its design with two types of roles: the chair and the experts. 

\begin{table}
\setlength{\tabcolsep}{9pt}
\caption{The impact of coordination mechanisms.}
\label{tab:mognet_ablation}
\centering
\begin{tabular}{@{}llllll@{}}
\toprule
                   & \bf BLEU    & \bf Inform  & \bf Success & \bf Score & \bf PPL  \\ 
\midrule
\acs{MoGNet}               & 20.13\% & 85.30\% & 73.30\% & 99.43 & 4.25 \\ 
\midrule 
\acs{MoGNet}-P & 19.51\% & 79.40\% & 71.80\% & 95.11 & 4.19 \\ 
\acs{MoGNet}-P-R         & \underline{18.16\%} & 85.10\% & 72.20\% & 96.81 & 4.12  \\
\bottomrule 
\end{tabular}
\begin{minipage}{\columnwidth}
\footnotesize{\underline{Underlined results} indicate the worst results with a statistically significant decrease compared to \acs{MoGNet} (paired t-test, $p < 0.01$).}
\end{minipage}
\end{table}

\subsection{Learning scheme}
\label{sec:analysis1}

We use \acs{MoGNet}-\acs{GL} to refer to the model that removes the \acs{GL} learning scheme from \acs{MoGNet} and uses the general global learning instead.
\acs{MoGNet}-\acs{GL} results in a sharp reduction of 6.95\% overall performance with 0.80\% of \textit{BLEU}, 6.90\% of \textit{Inform} and 5.40\% of \textit{Success}.
The main improvement is attributed to the strong task completion ability.
This shows the effectiveness and importance of the \acs{GL} learning scheme as it encourages each expert to specialize on a particular intent while the chair prompts all experts to coordinate with each other.

\begin{table}
\setlength{\tabcolsep}{9pt}
\caption{Impact of the learning scheme.}
\label{tab:mognet_learning_scheme}
\centering
\begin{tabular}{@{}llllll@{}}
\toprule
                   & \bf BLEU    & \bf Inform  & \bf Success & \bf Score & \bf PPL  \\ 
\midrule
\acs{MoGNet}               & 20.13\% & 85.30\% & 73.30\% & 99.43 & 4.25 \\ 
\acs{MoGNet}-\acs{GL}          & 19.33\% & \underline{78.40\%} & \underline{67.90\%} & \underline{92.48} & 3.97 \\ 
\bottomrule 
\end{tabular}
\begin{minipage}{\columnwidth}
\footnotesize{\underline{Underlined results} indicate the worst results with a statistically significant decrease compared with \acs{MoGNet} (paired t-test, $p < 0.01$).}
\end{minipage}
\end{table}


\section{ANALYSIS}

In this section, we explore \ac{MoGNet} in more detail. In particular, we examine 
\begin{enumerate*}[label=(\roman*)]
\item whether the intent partition affects the performance of \acs{MoGNet} (\S\ref{subsection:intent-partition-analysis});
\item whether the improvements of \acs{MoGNet} could simply be attributed to having a larger number of parameters (\S\ref{sec:analysis3});
\item how the hyper-parameter $\lambda$ (Eq.~\ref{finalloss}) affects the performance of \acs{MoGNet} (\S\ref{sec:analysis3}); and 
\item how \acs{RMoG}, \acs{PMoG} and \acs{GL} influence \ac{DRG} using a small case study (\S\ref{subsection:case-study}).
\end{enumerate*}

\subsection{Intent partition analysis}
\label{subsection:intent-partition-analysis}

As stated above, the responses vary a lot for different intents which are differentiated by the domain and the type of system action.
Therefore, we experiment with two types of intents as shown in Table~\ref{tab:intents}.

\begin{table}
\centering
\setlength{\tabcolsep}{1pt}
\caption{Two groups of intents that are divided by domains and the type of system actions.}
\label{tab:intents}
\begin{tabular}{ll}
\toprule
\bf Type & \bf Intents \\ 
\midrule
Domain      & \begin{tabular}[c]{@{}l@{}} Attraction, Booking, Hotel, Restaurant, Taxi, Train, General, UNK.\end{tabular} \\ 
\midrule
Action      & \begin{tabular}[c]{@{}l@{}} Book, Inform, NoBook, NoOffer, OfferBook, OfferBooked, Select, \\Recommend, Request,  Bye, Greet, Reqmore, Welcome, UNK.\end{tabular} \\ 
\bottomrule
\end{tabular}
\end{table}

\noindent%
To address (i), we compared two ways of partitioning intents.
\acs{MoGNet}-domain and \acs{MoGNet}-action denote the intent partitions w.r.t. domains and system actions, respectively.
\acs{MoGNet}-domain has 8 intents (domains) and \acs{MoGNet}-action has 14 intents (actions), as shown in Table~\ref{tab:intents}.
The results are shown in Table~\ref{tab:effect_on_intents}.
\begin{table}
\setlength{\tabcolsep}{8pt}
\caption{Results of \acs{MoGNet} with two intent partition ways. }
\label{tab:effect_on_intents}
\centering
\begin{tabular}{@{}lccccc@{}}
\toprule
                    & \bf BLEU    & \bf Inform  & \bf Success & \bf Score & \bf PPL  \\ \midrule
                         \acs{MoGNet}-domain         & \textbf{20.13\%} & \textbf{85.30}\% & \textbf{73.30}\% & \textbf{99.43} & \textbf{4.25} \\  
                         \acs{MoGNet}-action & 17.28\% & 79.40\% & 69.70\% & 91.83 & 4.48 \\ \bottomrule 
\end{tabular}
\end{table}

\acs{MoGNet} consistently outperforms the baseline \acs{S2SAttnGRU} for both ways of partitioning intents.
Interestingly, \acs{MoGNet}-domain greatly outperforms \acs{MoGNet}-action.
We believe there are two reasons:
First, the system actions are not suitable for grouping intents because some partition subsets are hard to be distinguished from each other, e.g., \textit{OfferBook} and \textit{OfferBooked}.
Second, some system actions only have a few data samples, simply not enough to specialize the experts.
The results show that different ways of partitioning intents may greatly affect the performance of \acs{MoGNet}.
Therefore, more effective intent partition methods, e.g., adaptive implicit intent partitions, need to be explored in future work.

\subsection{Hyper-parameter analysis}
\label{sec:analysis3}

\begin{figure}[htb!]
    \centering
    \includegraphics[clip,trim=0mm 2mm 0mm 4mm,width=\columnwidth]{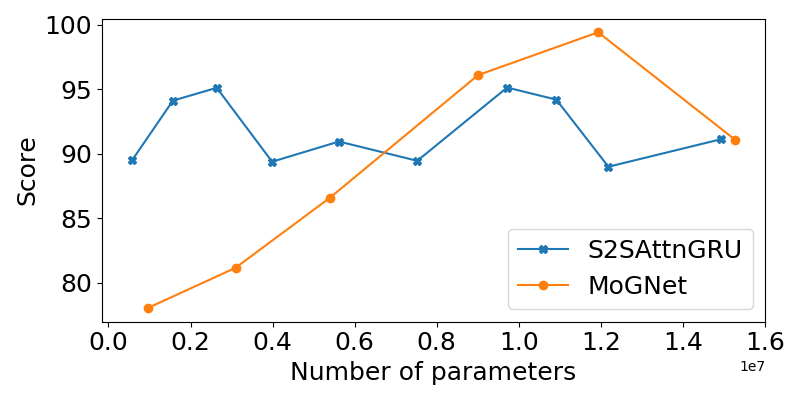}
    \caption{\textit{Score} of \acs{MoGNet} and \acs{S2SAttnGRU} with different number of parameters. }
    \label{fig:score_paranum}
\end{figure}

To address (ii), we show the results of \acs{MoGNet} and \acs{S2SAttnGRU} with different hidden sizes in Fig.~\ref{fig:score_paranum}.
\acs{S2SAttnGRU} outperforms \acs{MoGNet} when the number of parameters is less than 0.6e7.
However, \acs{MoGNet} achieves much better results with more parameters.
Most importantly, the results from both models show that a larger number of parameters does not always mean better performance, which indicates that the improvement of \acs{MoGNet} is not simply due to more parameters.

\begin{figure}[htb!]
    \centering
    \includegraphics[clip,trim=0mm 2mm 0mm 4mm,width=\columnwidth]{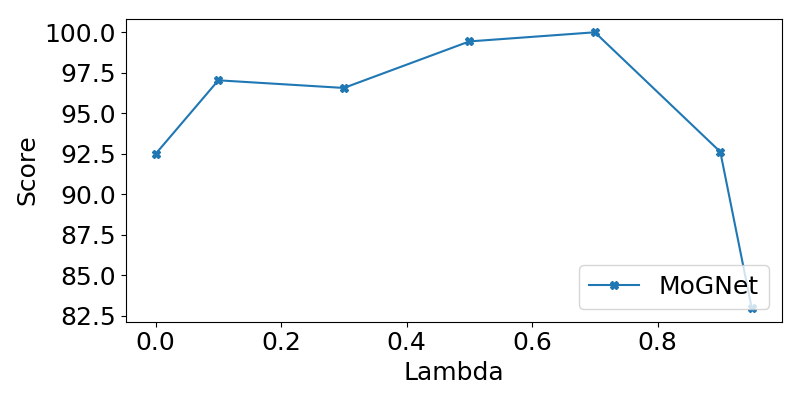}
    \caption{\textit{Score} of \acs{MoGNet} with different values of  $\lambda$.}
    \label{fig:score_lambda}
\end{figure}
%

To address (iii), we report the \textit{Score} values of \acs{MoGNet} with different values of $\lambda$ (Eq.~\ref{finalloss}), as shown in Fig.~\ref{fig:score_lambda}.
When $\lambda=0$, no expert is trained on a particular intent. 
When $\lambda=1$, the model ignores the global loss, i.e., the \acs{RMoG} and \acs{PMoG} mechanisms are not used and the chair is only trained as a general expert.
We can see that these two settings decrease the performance greatly which further verifies the effectiveness of \acs{RMoG} and \acs{PMoG} as well as the \acs{MoGNet} framework.
We also note that the performance of \acs{MoGNet} is quite stable when $\lambda \in [0.1, 0.7]$ with the best performance for $\lambda=0.7$.
Hence, \acs{MoGNet} is not very sensitive to the hyper-parameter $\lambda$.

\subsection{Case study}
\label{subsection:case-study}

\begin{table*}[t]
\caption{Example responses from \acs{MoGNet} with the ablation settings in a 4-turn dialogue.}
\label{tab:case_study}
\small
\centering
\setlength{\tabcolsep}{1pt}
\begin{tabular}{@{}cll@{}}
\toprule
\normalsize \bf Turn & \normalsize \bf User & \normalsize \bf System \\ \midrule
1    & \begin{tabular}[c]{@{}l@{}}i am looking for a train departing \\ from {[}value\_place{]} that will arrive \\ at {[}value\_place{]}.\end{tabular} & \begin{tabular}[c]{@{}l@{}}
\textbf{\acs{S2SAttnGRU}}: what {\color{red}day} would you like to travel ?\\
\textbf{\acs{LaRLAttnGRU}}: there are {[}value\_count{]} trains leaving {[}value\_place{]} on {[}value\_day{]} . what\\ {\color{red}day} would you like to travel ?\\
\midrule
\textbf{MoGNet}: what {\color{red}day} and {\color{red}time} would you like to travel?\\ 
\textbf{MoGNet-P}: what {\color{red}day} will you be traveling?\\ 
\textbf{MoGNet-P-R}: what {\color{red}day} will you be traveling?\\ 
\textbf{MoGNet-GL}: there are {[}value\_count{]} trains ... do you have a {\color{red}time} you would like to depart?
\end{tabular}\\ \midrule
2    & \begin{tabular}[c]{@{}l@{}}i would like to leave on {[}value\_day{]} \\ and arrive no later than {[}value\_time{]}.\end{tabular}                 & \begin{tabular}[c]{@{}l@{}}
\textbf{\acs{S2SAttnGRU}}: {[}train\_id{]} will arrive at {\color{red}{[}value\_time{]}} .\\
\textbf{\acs{LaRLAttnGRU}}: i have train {[}train\_id{]} leaving at {\color{red}{[}value\_time{]}} and arriving at {[}value\_time{]} . \\would you like me to book this for you ?\\
\midrule
\textbf{MoGNet}: {[}train\_id{]} leaves at {\color{red}{[}value\_time{]}} and arrives by {[}value\_time{]} .\\ 
\textbf{MoGNet-P}: {[}train\_id{]} will arrive at {[}value\_time{]}.\\ 
\textbf{MoGNet-P-R}: the {[}train\_id{]} arrives at {[}value\_time{]}.\\ 
\textbf{MoGNet-GL}: i have train {[}train\_id{]} that leaves at {\color{red}{[}value\_time{]}} and arrives at {[}value\_time{]}. \\ +\textit{All}: would you like me to book that for you?
\end{tabular}\\ \midrule
3    & can i please have the reference number?& \textit{All}: ... how many tickets do you need?\\ \midrule
4    & \begin{tabular}[c]{@{}l@{}}could you book {[}value\_count{]} tickets \\ for me and send me the reference\\ number?\end{tabular}                    & \begin{tabular}[c]{@{}l@{}}
\textbf{\acs{S2SAttnGRU}}: i have booked you {[}value\_count{]} tickets on the {[}train\_id{]} . your reference\\ number is {[}train\_reference{]}. is there anything else i can help you with ?\\
\textbf{\acs{LaRLAttnGRU}}: i have booked {[}value\_count{]} tickets on {[}train\_id{]} . the reference number is \\ {[}train\_reference{]} .\\
\midrule
\textbf{MoGNet}: i have booked {[}value\_count{]} tickets for you. the total fee is {\color{red}{[}value\_price{]}} gbp\\ payable at the station. your reference number is {[}train\_reference{]}. is there anything else\\ i can help you with ?\\ 
\textbf{MoGNet-P}: i have booked {[}value\_count{]} tickets. the total fee is {\color{red}{[}value\_price{]}} gbp payable\\ at the station. your reference number is {[}train\_reference{]}.\\ 
\textbf{MoGNet-P-R}: booking was successful , the total fee is {\color{red}{[}value\_price{]}} gbp payable at the\\ station. reference number is {[}train\_reference{]}.\\ 
\textbf{MoGNet-GL}: i have booked {[}value\_count{]} tickets for you. the reference number is\\ {[}train\_reference{]}. is there anything else i can help you with ?
\end{tabular} \\ \bottomrule
\end{tabular}
\end{table*}
\vspace*{-0.3\baselineskip}

To address (iv), we select an example to illustrate the influence of \acs{RMoG}, \acs{PMoG}, and \acs{GL}.
Table~\ref{tab:case_study} exhibits the responses generated by comparable baselines (i.e., \acs{S2SAttnGRU}, \acs{LaRLAttnGRU}) and \acs{MoGNet} variants as in Table~\ref{tab:mognet_ablation}.
In red we highlight the tokens that show the differences in terms of task completion.
Generally, \acs{MoGNet} can generate more appropriate and meaningful responses.
Specifically, without \acs{PMoG}, \acs{MoGNet}-P and \acs{MoGNet}-P-R ignore the fact that the attribute \textit{time} is important for searching a train ticket (1st turn) and omit the exact departure time ({[value\_time]}) of the train (2nd turn).
Without \acs{GL}, \acs{MoGNet}-GL ignores the primary time information need \textit{day} (1st turn) and omits the implicit need of {[value\_price]} (4th turn).
There are also some low-quality cases, e.g., \acs{MoGNet} and the baselines occasionally generate redundant and lengthy responses, because none of them has addressed this issue explicitly during training.


\section{RELATED WORK}

Traditional models for \acs{DRG}~\cite{crook2016tds_platform,yan2017building} decompose the task into sequentially dependent modules, e.g., \ac{DST}~\cite{zhong2018dst}, \ac{PL}~\cite{zhang2019memory}, and \ac{NLG}~\cite{mi2019meta}.
Such models allow for targeted failure analyses, but inevitably incur upstream propagation problems~\cite{chen2017survey}. 
Recent work views \acs{DRG} as a source-to-target transduction problem, which maps a \textit{dialogue context} to a \textit{response}~\cite{eric2017key,li2017adversarial,wen2017network}.
\citet{sordoni2015neural} show that using an \ac{RNN} to generate text conditioned on dialogue history results in more natural conversations. 
Later improvements include the addition of attention mechanisms~\cite{li2016persona,vinyals2015neural}, modeling the hierarchical structure of dialogues~\cite{serban2016building}, or jointly learning belief spans~\cite{lei2018sequicity}.
Strengths of these methods include global optimization and easier adaptation to new domains~\cite{chen2017survey}.

The studies listed above assume that each token of a response is sampled from a single distribution, given a complex dialogue context.
In contrast, \acs{MoGNet} uses multiple cooperating modules, which exploits the specialization capabilities of different experts and the generalization capability of a chair.
Work most closely related to ours in terms of modeling multiple experts includes
\citep{chen2019semantically,guo2018multi,le2016lstm,pei2019modular}.
\citet{le2016lstm} integrate a chat model with a question answering model using an LSTM-based mixture-of-experts method. Their model is similar to \acs{MoGNet}-GL-P (without \ac{PMoG} and \ac{GL}) except that they simply use two implicit expert generators that are not specialized on particular intents.
\citet{guo2018multi} introduce a mixture-of-experts to use the data relationship between multiple domains for binary classification and sequence tagging. 
Sequence tagging generates a set of fixed labels; \acs{DRG} generates diverse appropriate response sequence.
The differences between \acs{MoGNet} and these two approaches are three-fold:
First, \acs{MoGNet} consists of a group of modules including a chair generator and several expert generators; this design addresses the module interdependence problem since each module is independent from the others.
Second, the chair generator alleviates the error propagation problem because it is able to manage the overall errors through an effective learning scheme.
Third, the models of those two approaches cannot be directly applied to task-oriented \acs{DRG}.
The recently published HDSA~\cite{chen2019semantically} slightly outperforms \acs{MoGNet} on \textit{Score} (+0.07), but it overly relies on BERT~\cite{devlin2019bert} and graph structured dialog acts.
\acs{MoGNet} follow the same modular \acs{TDS} framework~\cite{pei2019modular}, but it preforms substantially better due to fitting the expert generators with both retrospection and prospection abilities and adopting the \ac{GL} learning scheme to conduct more effective learning.


\section{CONCLUSION AND FUTURE WORK}
\label{sec:conclusion}

In this paper, we propose a novel \acf{MoGNet} model with different coordination mechanisms, namdely, \ac{RMoG} and \ac{PMoG}, to enhance \acl{DRG}.
We also devise a \ac{GL} learning scheme to effectively learn \acs{MoGNet}.
Experiments on the MultiWOZ benchmark demonstrate that \acs{MoGNet} significantly outperforms state-of-the-art methods in terms of both automatic and human evaluations.
We also conduct analyses that confirm the effectiveness of \acs{MoGNet}, the \ac{RMoG} and \ac{PMoG} mechanisms, as well as the \ac{GL} learning scheme.

As to future work, we plan to devise more fine-grained expert generators and to experiment on more datasets to test \acs{MoGNet}.
In addition, \acs{MoGNet} can be advanced in many directions:
First, better mechanisms can be proposed to improve the coordination between chair and expert generators.
Second, it would be interesting to study how to do intent partition automatically.
Third, it is also important to investigate how to avoid redundant and lengthy responses in order to provide a better user experience.

\ack 
This research was partially supported by
Ahold Delhaize,
the Association of Universities in the Netherlands (VSNU),
the China Scholarship Council (CSC),
and
the Innovation Center for Artificial Intelligence (ICAI).
All content represents the opinion of the authors, which is not necessarily shared or endorsed by their respective employers and/or sponsors.

\bibliographystyle{abbrvnat}
\bibliography{ecai}

\end{document}